\documentclass[letterpaper, 10 pt, conference]{ieeeconf}  

\IEEEoverridecommandlockouts                              

\usepackage{amsmath} 
\usepackage{amssymb}  
\usepackage{dsfont}
\usepackage{graphicx}

\usepackage{psfrag,graphicx,epsfig}
\usepackage{epstopdf}
\usepackage{xspace}
\usepackage{subfig}
\usepackage{float}
\usepackage{placeins}
\usepackage{multirow}
\usepackage{pgf,tikz}
\usepackage{nowidow}
\usepackage{lineno}
\usepackage{xcolor}
\newcommand{\fig}[1]{Fig.~\ref{#1}}
\newcommand{\tab}[1]{Table~\ref{#1}}
\newcommand{\eq}[1]{(\ref{#1})}

\DeclareMathOperator*{\argmax}{arg\,max}
\usepackage{siunitx}
\usepackage{color}
\usepackage{flushend}
\usepackage{lineno}
\usepackage{algorithmic}
\allowdisplaybreaks

\title{\LARGE \bf
Robust Pivoting: Exploiting Frictional Stability\\ Using Bilevel Optimization
}

\author{Yuki Shirai$^{\dagger}$, Devesh K. Jha$^{\ddagger}$, Arvind U. Raghunathan$^{\ddagger}$ and Diego Romeres$^{\ddagger}$%
\thanks{$^{\dagger}$ Yuki Shirai is with the Department of Mechanical and Aerospace Engineering, University of California, Los Angeles, CA, USA 90095 {\tt\small yukishirai4869@g.ucla.edu}}%
\thanks{$^{\ddagger}$Devesh K. Jha, Arvind U. Raghunathan and Diego Romeres are with Mitsubishi Electric Research Laboratories (MERL), Cambridge, MA, USA 02139 {\tt\small \{jha,raghunathan,romeres\}@merl.com}}}%

\begin{document}

\maketitle

\begin{abstract}
Generalizable manipulation requires that robots be able to interact with novel objects and environment. This requirement makes manipulation extremely challenging as a robot has to reason about complex frictional interaction with uncertainty in physical properties of the object. In this paper, we study robust optimization for control of pivoting manipulation in the presence of uncertainties. We present insights about how friction can be exploited to compensate for the inaccuracies in the estimates of the physical properties during manipulation. In particular, we derive analytical expressions for stability margin provided by friction during pivoting manipulation. This margin is then used in a bilevel trajectory optimization algorithm to design a controller that maximizes this stability margin to provide robustness against uncertainty in physical properties of the object. We demonstrate our proposed method using a 6 DoF manipulator for manipulating several different objects. 
\end{abstract}
\section{Introduction}\label{sec:introduction}
Contacts are central to most manipulation tasks as they provide additional dexterity to robots to interact with their environment~\cite{mason2018toward}.  Designing robust controllers for frictional interaction with objects with uncertain physical properties is challenging as the mechanical stability of the object depends on these physical properties. Inspired by this problem, we consider the task of pivoting manipulation in this paper. In particular, we consider the problem of re-orienting parts with uncertain mass and Center of Mass (CoM) location using pivoting. We are interested in ensuring mechanical stability via friction to compensate for uncertainty in the physical properties of the objects.

\begin{figure}
    \centering
    \includegraphics[width=0.49\textwidth]{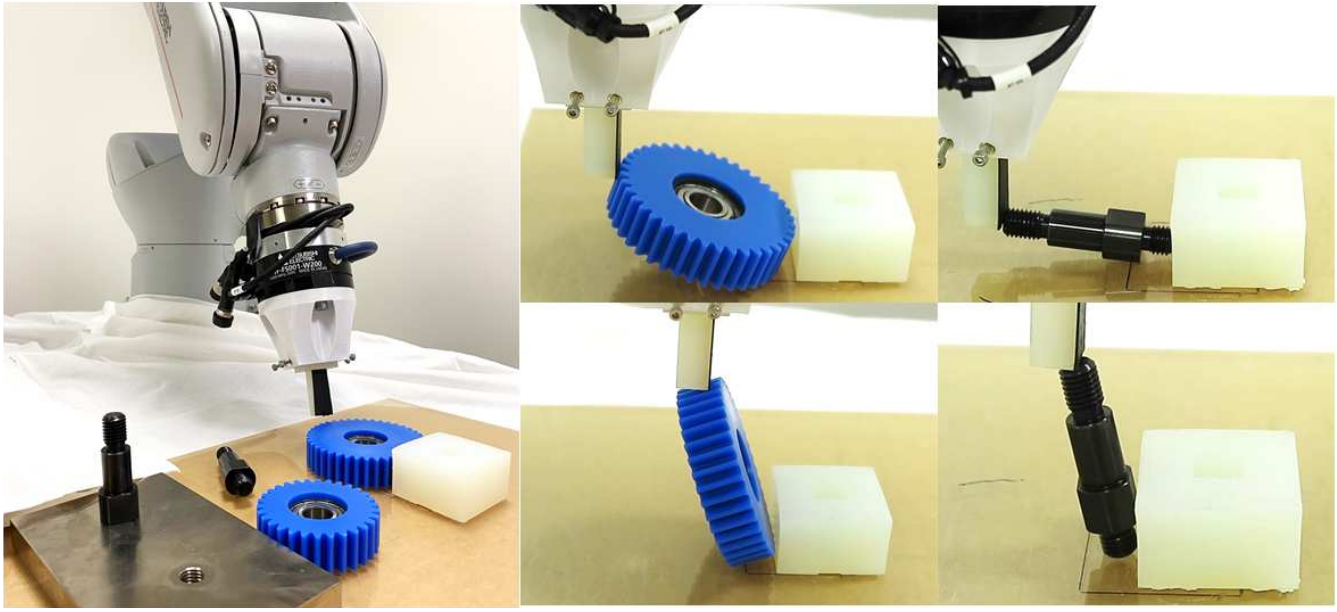} %
    \caption{We consider the problem of reorienting parts for assembly using pivoting manipulation primitive. Such reorientation could possibly be required when the parts being assembled are too big to grasp in the initial pose (such as the gears) or the parts to be inserted during assembly are not in the desired pose (such as the pegs). The figure shows some instances during the implementation of our  controller to reorient a gear and a peg.}
    \label{fig:pivoting_abstractfig}
\end{figure}
Designing robust controllers for frictional interaction systems is challenging due to the  hybrid nature of underlying frictional dynamics. Consequently, a lot of classical robust planning and control techniques are not applicable to these systems in the presence of  uncertainties~\cite{drnach2021robust,DBLP:journals/corr/abs-2105-09973}.
While concepts of stability margin or Lyapunov stability have been well studied in the context of nonlinear dynamical system controller design~\cite{vidyasagar2002nonlinear}, such notions have not been explored in contact-rich manipulation problems. This can be mostly attributed to the fact that a controller has to reason about the mechanical stability constraints of the frictional interaction to ensure stability. The mechanical stability closely depends on the contact configuration during manipulation, and thus a controller has to ensure that the desired contact configuration is either maintained during the task or it can maintain stability even if the contact sequence is perturbed. Analysis of such systems is difficult in the presence of friction as it leads to a differential inclusion system (see~\cite{raghunathan2020stability}) . One of the key insights we present in this paper is that friction provides mechanical stability margin during a contact-rich task. We call the mechanical stability provided by friction as \textit{Frictional Stability}. This \textit{frictional stability} can be exploited during optimization to allow stability of manipulation in the presence of uncertainty. 

We study pivoting manipulation where the object being manipulated has to maintain slipping contact with two external surfaces (see \fig{fig:mechanics_pivoting_eq}). A robot can use this manipulation to reorient parts on a planar surface to allow grasping or assist in assembly by manipulating objects to a desired pose (see \fig{fig:pivoting_abstractfig}). Note that this manipulation is challenging as it requires controlled slipping (as opposed to sticking contact~\cite{hou2018fast, hogan2020tactile}), and thus it is imperative to consider robustness of the control trajectories. To ensure mechanical stability of the two-point pivoting in the presence of uncertainty, we derive a sufficient condition for stability which allows us to compute a margin of stability. This margin is then used in a bilevel optimization routine to design an optimal control trajectory while maximizing this margin.



\textbf{Contributions.} This paper has the following contributions.
\begin{enumerate}
    \item We present analysis of mechanical stability of pivoting with uncertainty in mass and CoM location of objects.
    \item We present a bilevel optimization technique which can be used to optimize the mechanical stability margin during the pivoting manipulation.
    \item The proposed method is demonstrated for reorienting parts using a 6 DoF manipulator.
\end{enumerate}
\section{Related Work}\label{sec:related_work}
Contact modeling has been extensively studied in mechanics as well as robotics literature~\cite{todorov2010implicit, drumwright2011evaluation, 9366782}. One of the most common contact models is based on the linear complementarity problem (LCP). LCP-based contacts models have been extensively used for performing trajectory optimization in manipulation~\cite{DBLP:journals/corr/abs-2106-03220,jin2021trajectory} as well as locomotion~\cite{posa2014direct}. More recently, there has also been some work for designing robust manipulation techniques for contact-rich systems using stochastic optimization~\cite{drnach2021robust, DBLP:journals/corr/abs-2105-09973}. 
These problems consider stochastic complementarity systems and consider robust optimization for the underlying stochastic system. However, these problems consider a dynamical model and do not explicitly consider the mechanical stability during planning. Our work is motivated by the concepts of stability under multiple contacts in legged locomotion. Static stability with multiple contacts has been widely studied in legged locomotion~\cite{4598894, 8383993, 8416785, 8358969, 9113247}. These works consider the problem of mechanical stability of the legged robot under multiple contacts by considering the stability polygon defined by the frictional contacts. Similar to the concept of stability polytope, we present the idea of frictional stability which defines the extent to which multiple points of contact can compensate for gravitational force and moments in the presence of uncertainty in the mass and CoM location.

In~\cite{hogan2020tactile}, authors consider stabilization of a table-top manipulation task during online control. They consider a decomposition of the control task in object state control and contact state control. The contact state was detected using vision-based tactile sensors~\cite{donlon2018gelslim,li2020f, dong2021icra}. As the task mostly required sticking contact for stability, the tactile feedback was designed to make corrections to push the system away from the boundary of friction cone at the different contact locations. However, the authors did not consider the problem of designing trajectories which can provide robustness to uncertainty. Furthermore, the authors only considered controlled sticking in~\cite{hogan2020tactile} which is, in general, easier than controlled slipping.  Other previous works that study stable pivoting also consider sticking contact during pivoting using multiple points of contact~\cite{hou2018fast}. The problem in~\cite{hou2018fast} is inherently stable as the object is always in stable grasp. Furthermore, the authors do not consider any uncertainty during planning. Similarly, authors in~\cite{aceituno2020global} present a mixed integer programming formulation to generate contact trajectory given a desired reference trajectory for the object for several manipulation primitives. 
Another related work is presented in~\cite{han2020local} where the authors study the feedback control during manipulation of a half-cylinder. The idea there is to design a reference trajectory and then use a local controller by building a funnel around the reference trajectory by linearizing the dynamics. The online control is computed by solving linear programs to locally track the reference trajectory.


\section{Mechanics of Pivoting}\label{sec:mechanics}

\begin{figure}
    \centering
    \includegraphics[width=0.255\textwidth]{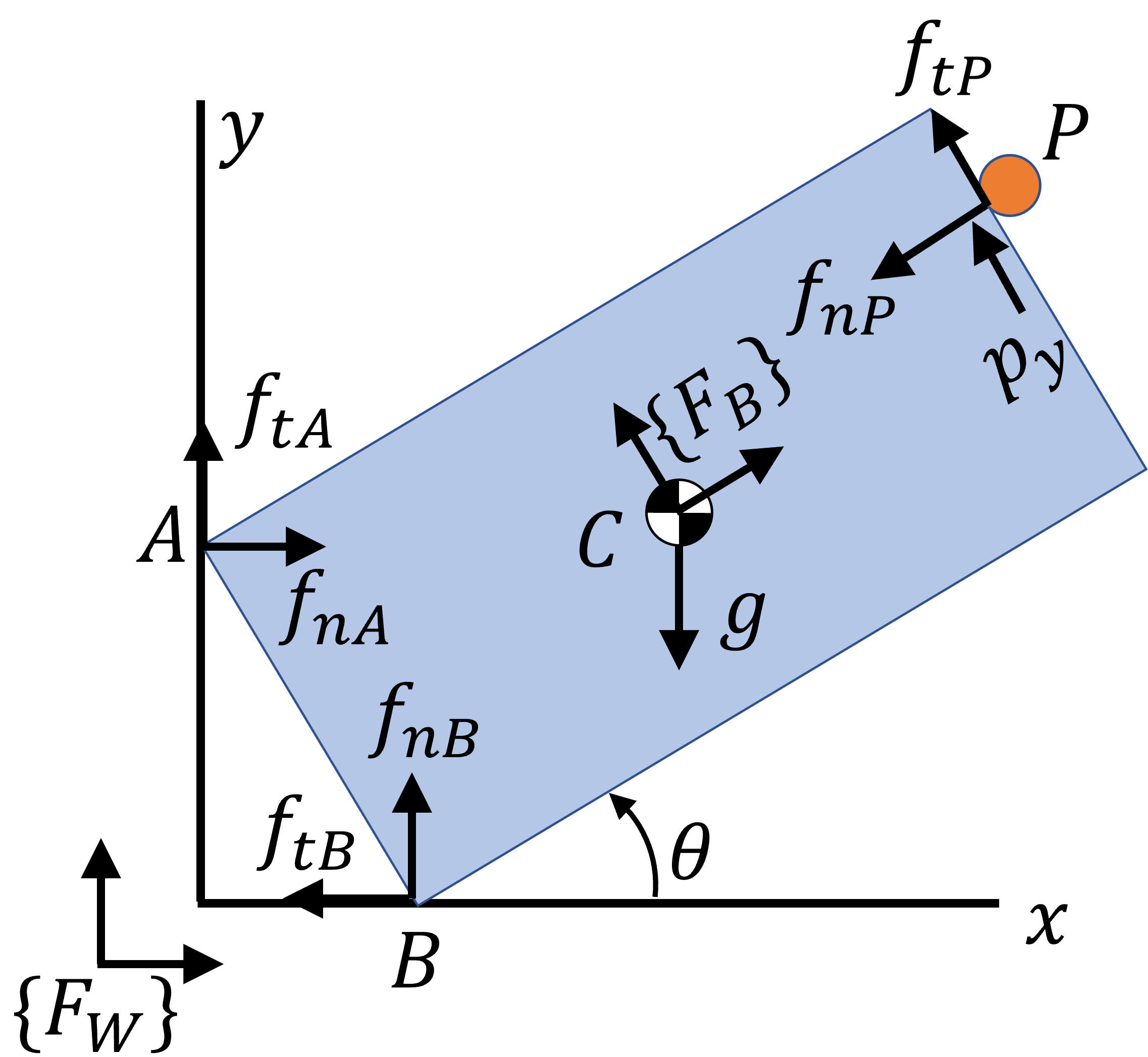} %
    \caption{A schematic showing the free-body diagram of a rigid body during pivoting manipulation. Point $P$ is the contact point with a manipulator.}
    \label{fig:mechanics_pivoting_eq}
\end{figure}

In this section, we explain quasi-static stability of two-point pivoting in a plane. 
Before explaining the details, we present our assumptions in this work:
\begin{enumerate}
\item The object is rigid.
\item We consider static equilibrium of the object. 
\item The external contact surfaces are perfectly flat. 
\item The dimensions of the object and the frictional parameters are perfectly known.
\end{enumerate}

\subsection{Mechanics of Pivoting with External Contacts}
We consider pivoting where the object maintains slipping contact with two external surfaces (see \fig{fig:mechanics_pivoting_eq}). A free body diagram showing the static equilibrium of the object is shown in \fig{fig:mechanics_pivoting_eq}. The object experiences four forces corresponding to two friction forces $f_A, f_B$ from external contact points $A$ and $B$, one control input $f_P$ from manipulator at point $P$, and gravity, $mg$ at point $C$ where $m$ is mass of a body. We denote $f_{ni}, f_{ti}$ as a normal force and friction force at point $\forall i, i=\{A, B\}$, respectively, defined in ${\{F_W\}}$. $f_{nP}, f_{tP}$ are normal and friction force at point $P$ defined in ${\{F_B\}}$. Note that we define the $[f_x, f_y]^\top = \mathbf{R} [f_{nP}, f_{tP}]^\top$ where $\mathbf{R}$ is a rotation matrix from ${\{F_B\}}$ to ${\{F_W\}}$. We denote $x, y$ position at point in ${\{F_W\}}$ $\forall i, i=\{A, B, P\}$ as $i_x, i_y$, respectively. We denote $y$ position of point $P$ in ${\{F_B\}}$ as $p_y$.  We define the angle of body with respect to $x$-axis as $\theta$. The coefficient of friction at point $\forall i, i=\{A, B, P\}$ are $\mu_A, \mu_B, \mu_P$, respectively. 
In the later sections we present trajectory optimization formulation where we consider friction force variables $f_{ni}, f_{ti}$, contact point variables $i_{x}, i_{y}$ $\forall i, i=\{A, B, P\}$, $\theta$, and $p_y$ at each time-step $k$ denoted as $f_{k, ni}, f_{k, ti}, i_{k, x}, i_{k, y}, \theta_k, p_{y, k}$. In this section, we remove $k$ to represent variables for simplicity. 

The static equilibrium conditions for the object can be represented by the following equations (note we consider the moment at point $B$ by setting $B_x = B_y = 0$):
\begin{subequations}
\begin{flalign}
 f_{nA} + f_{tB} + f_{xP}  =0,\label{forceeq1}\\
f_{tA} + f_{nB} + mg + f_{yP}   = 0,  \label{forceeq2}\\
A_xf_{tA} - A_yf_{nA} + C_xmg + P_xf_{y} - P_y f_x = 0 \label{moment_eq1}
\end{flalign}
\label{force_eq}
\end{subequations}
We consider Coulomb friction law which results in friction cone constraints as follows:
\begin{equation}
 |f_{tA}|  \leq \mu_A f_{nA}, |f_{tB}|  \leq \mu_B f_{nB}, \quad f_{nA}, f_{nB} \geq 0
\label{general_FC}
\end{equation}
To describe sticking-slipping complementarity constraints, we have the following complementarity constraints at point $i = \{A, B\}$:
\begin{subequations}
\begin{flalign}
 0 \leq   \dot{p}_{i+} \perp \mu_i f_{ni}-f_{ti} \geq 0  \\
 0 \leq   \dot{p}_{i-} \perp \mu_i  f_{ni}+f_{ti} \geq 0 
 \end{flalign}
 \label{slippingAB}
\end{subequations}
where the slipping velocity at point $i$ follows $\dot{p}_i=\dot{p}_{i+}-\dot{p}_{i-}$.
$\dot{p}_{i+}, \dot{p}_{i-}$ represent the slipping velocity along positive and negative directions for each axis, respectively.
Since we consider slipping contact during pivoting, we have "equality" constraints in friction cone constraints at points $A, B$:
\begin{equation}
 f_{tA}  =\mu_A f_{nA}, f_{tB}  =-\mu_B f_{nB}
\label{slipping_friction_cone}
\end{equation}
To realize stable pivoting, actively controlling position of point $P$ is important. Thus, we consider the following complementarity constraints that represent the relation between the slipping velocity $\dot{p}_y$  at point $P$ and friction cone constraint at point $P$:
\begin{subequations}
\begin{flalign}
 0 \leq   \dot{p}_{y+} \perp \mu_p f_{nP}-f_{tP} \geq 0  \\
 0 \leq   \dot{p}_{y-} \perp \mu_p  f_{nP}+f_{tP} \geq 0 
 \end{flalign}
 \label{slippingP}
\end{subequations}
where $\dot{p}_y=\dot{p}_{y+}-\dot{p}_{y-}$. 



\subsection{Frictional Stability Margin}


We briefly provide some physical intuition about frictional stability which is formalized in the later sections. First, suppose that uncertainty exists in mass of a body. In the case when the actual mass is lower than  estimated, the friction force at point $A$ would increase while the friction force at point $B$ would decrease, compared to the nominal case. In contrast, suppose if the actual mass of the body is heavier than that of what we estimate, then the body can tumble along point $B$ in the clockwise direction. In this case, we can imagine that the friction force at point $A$ would decrease while the friction force at point $B$ would increase. However, as long as the friction forces are non-zero, the object can stay in contact with the external environment.
Similar arguments could be made for uncertainty in CoM location. The key point to note that the friction forces can re-distribute at the two contact locations and thus provide a margin of stability to compensate for uncertain gravitational forces and moments. We call this margin as \textit{frictional stability}.



\subsection{Stability Margin for Uncertain Mass}\label{sec_uncertain_mass}
 For simplicity, we denote $\epsilon$ as uncertain weight with respect to the estimated weight. Also, to emphasize that we consider the system under uncertainty, we put superscript $\epsilon$ for each friction force variable. Thus, the static equilibrium conditions in \eq{force_eq} can be rewritten as:
\begin{subequations}
\begin{flalign}
f_{nA}^\epsilon + f_{tB}^\epsilon + f_{xP}  =0\label{forceeq11},\\
f_{tA}^\epsilon + f_{nB}^\epsilon + (mg + \epsilon) + f_{yP}   = 0,  \label{forceeq21}\\
A_xf_{tA}^\epsilon - A_yf_{nA}^\epsilon + C_x (mg+\epsilon) + P_xf_{y} - P_y f_x = 0 \label{moment_eq11}
\end{flalign}
\label{force_eq_mass}
\end{subequations}
Then, using \eq{slipping_friction_cone} and \eq{moment_eq11}, we obtain:
\begin{equation}
f_{nA}^\epsilon = \frac{-{C_x}\left(mg + \epsilon\right) -{P_x}f_{y} + {P_y}f_{x}}{\mu_A {A_x} - {A_y}}
\label{fna_111}
\end{equation}
To ensure that the body maintains contact with the external surfaces, we would like to enforce that the body experiences non-zero normal forces at both contacts.
To realize this, we have $f_{nA}^\epsilon \geq 0, f_{nB}^\epsilon \geq 0$ as conditions that the system needs to satisfy. Consequently, by simplifying \eq{fna_111}, we get the following:
\begin{subequations}
\begin{flalign}
\epsilon \geq \frac{P_yf_x - P_xf_y - C_xmg}{C_x}, \text{ if } C_x>0,  \label{fnacond1}\\
\epsilon \leq \frac{P_yf_x - P_xf_y - C_xmg}{C_x}, \text{ if } C_x<0  \label{fnacond2}
\end{flalign}
\label{fna_cond_mass}
\end{subequations}
Note that the upper-bound of $\epsilon$ means that the friction forces can exist even when we make the mass of the body lighter up to $\frac{\epsilon}{g}$. The lower-bound of $\epsilon$ means that the friction forces can exist even when we make the mass of the body heavier up to $\frac{\epsilon}{g}$. 
\eq{fna_cond_mass} provides some useful insights. \eq{fna_cond_mass} gives either upper- or lower-bound of $\epsilon$ for $f_{nA}^\epsilon$ according to the sign of $C_x$ (the moment arm of gravity). This is because the uncertain mass would generate an additional moment along with point $B$ in the clock-wise direction if $C_x >0$ and in the counter clock-wise direction if $C_x <0$. 
If $C_x = 0$, we have an unbounded range for $\epsilon$, meaning that the body would not lose contact at point $A$ no matter how much uncertainty exists in the mass. 

\eq{fna_cond_mass} can be reformulated as an inequality constraint: 
\begin{equation}
 C_x(\epsilon - \epsilon_A) \geq 0
\label{fna_cond_mass_one}
\end{equation}
where $\epsilon_A = \frac{P_yf_x - P_xf_y - C_xmg}{C_x}$.

We can derive condition for $\epsilon$ based on $f_{nB}^\epsilon \geq 0$ from \eq{slipping_friction_cone}, \eq{forceeq11}, and \eq{forceeq21}:
\begin{equation}
 \epsilon \leq \mu_A f_x -f_y -mg
\label{fnb_cond_mass}
\end{equation}
We only have upper-bound on $\epsilon$ based on $f_{nB}^\epsilon \geq 0$, meaning that the contact at point $B$ cannot be guaranteed if the actual mass is lighter than $\mu_A f_x -f_y -mg$. 

\subsection{Stability Margin for Uncertain CoM Location}
We denote $dx, dy$ as residual CoM locations with respect to the estimated CoM location in ${\{F_B\}}$ coordinate, respectively. Thus, the residual CoM location in $x_W, y_W$, $dx_W, dy_W$, are represented by $dx_W = d \cos({\theta + \theta_d}), dy_W = d \sin({\theta + \theta_d})$, where $d = \sqrt{dx^2 + dy^2}$, $\theta_d = \arctan{\frac{dy}{dx}}$.  For notation simplicity, we use $r$ to represent $dx_W$.  In this paper, we put superscript $r$ for each friction force variables. Then, the static equilibrium conditions in \eq{force_eq} can be rewritten as follows:
\begin{subequations}
\begin{flalign}
f_{nA}^r + f_{tB}^r + f_{xP}  =0\label{forceeq12},\\
f_{tA}^r + f_{nB}^r + mg + f_{yP}   = 0,  \label{forceeq22}\\
A_xf_{tA}^r - A_yf_{nA}^r + (C_x + r) mg + P_xf_{y}  = P_y f_x  \label{moment_eq12}
\end{flalign}
\label{force_eq_location}
\end{subequations}
Then, using \eq{slipping_friction_cone} in \eq{force_eq_location}, we obtain:
\begin{subequations}
\begin{flalign}
r \leq \frac{P_yf_x -P_x f_y}{mg} - C_x \label{fna_fnb_r1},\\
r \geq -C_x - \frac{\frac{\mu_A A_x - A_y}{1 + \mu_A}(-f_x-f_y-mg) -P_yf_x + P_xf_y}{mg}  \label{fna_fnb_r2}
\end{flalign}
\label{fna_fnb_r}
\end{subequations}
where \eq{fna_fnb_r1}, \eq{fna_fnb_r2} are obtained based on $f_{nA}^r \geq 0, f_{nB}^r \geq 0$, respectively. \eq{fna_fnb_r} means that the object would lose contact at $A$ if the actual CoM location is more to the right than our expected CoM location while the object would lose the contact at $B$ if the actual CoM location is more to the left.

\section{Robust Trajectory Optimization}\label{sec:robust_to}
\begin{figure}
    \centering
    \includegraphics[width=0.45\textwidth]{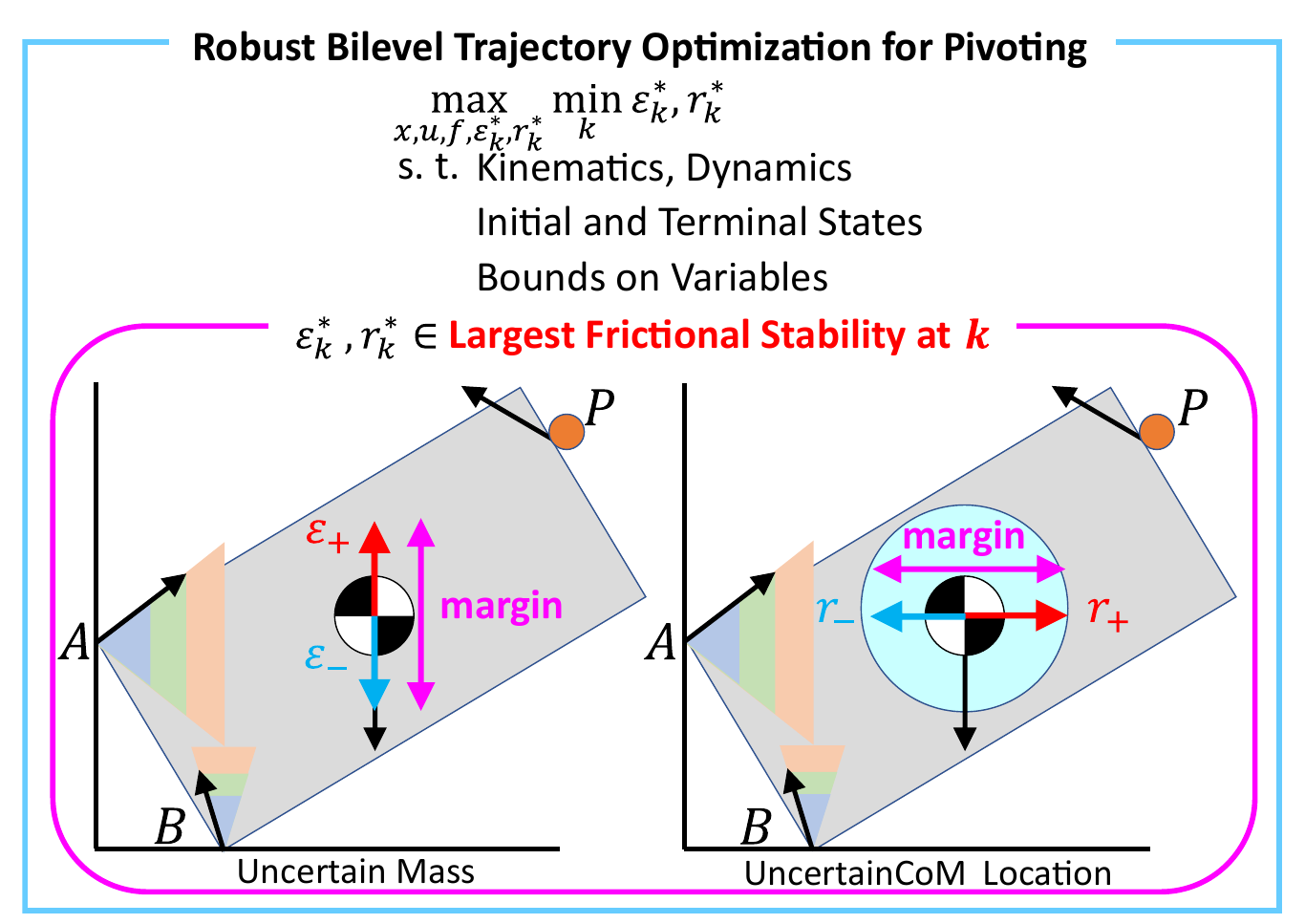} %
    \caption{Conceptual schematic of our proposed frictional stability and robust trajectory optimization for pivoting. Due to slipping contact, friction forces at points $A, B$ lie on the edge of friction cone. Given the nominal trajectory of state and control inputs, friction forces can account for uncertain physical parameters to satisfy static equilibrium. We define the range of uncertainty in gravitational forces and moments that can be compensated by contacts as frictional stability.}
    \label{fig:concept}
\end{figure}
In this section, we describe our robust trajectory optimization based on bilevel optimization which explicitly considers frictional stability under uncertainty of mass and CoM location. Our proposed method is also presented as a schematic in \fig{fig:concept}.
\subsection{Contact-Implicit Trajectory Optimization for Pivoting}


The purpose of our optimal control is to regulate the contact state and object state simultaneously given by:
%
\begin{subequations}
\begin{flalign}
\min _{x, u, f} \sum_{k=0}^{N-1} ({x}_{k} - x_g)^{\top} Q ({x}_{k} - x_g)+u_{k}^{\top} R u_{k} \\
\text{s. t. } i_{k, x}, i_{k, y} \in FK(x_k), \eq{force_eq}, \eq{slipping_friction_cone}, \eq{slippingP},  \label{const2}\\
x_{0} = x_s, x_{N} = x_g,
x_{k} \in \mathcal{X}, u_{k} \in \mathcal{U}, 0\leq f_{k, ni} \leq f_{u} \label{bounds_variables}
\end{flalign}
\label{equation_control}
\end{subequations}
where $x_k = [\theta_k, p_{k, y}]^\top$, $u_k=[f_{k, nP}, f_{k, tP}]^\top$, $f_k = [f_{k, nA},f_{k, nB}]^\top$, $Q=Q^{\top} \geq 0,R=R^{\top} > 0$. The function $FK$ represents forward kinematics to specify each contact point $i$ and CoM location. $\mathcal{X}$ and $\mathcal{U}$ are convex polytopes, consisting of a finite number of linear inequality constraints.  $f_u$ is an upper-bound of normal force at each contact point. Note that we impose \eq{force_eq}, \eq{slipping_friction_cone} at each time step $k$. $x_s, x_g$ are the states at $k=0,k=N$, respectively.



\subsection{Robust Bilevel Contact-Implicit Trajectory Optimization }
In this section, we present our formulation where we incorporate frictional stability in trajectory optimization to obtain robustness. An important point to note is that the optimization problem would be ill-posed if we naively add \eq{force_eq_mass} and/or \eq{force_eq_location} to \eq{equation_control} since there is no $u$ to satisfy all uncertainty realization in equality constraints. Therefore, our strategy is that we plan to find an optimal nominal trajectory that can ensure external contacts under mass or CoM location uncertainty. In other words, we aim at maximizing the worst-case stability margin over the trajectory given the maximal frictional stability at each time-step $k$ (also shown in \fig{fig:concept}). Thus,  we maximize the following objective function:
\begin{equation}
\min _{k} \epsilon_{k, +}^* - \max _{k} -\epsilon_{k, -}^*
\label{bilevel_obj}
\end{equation}
where $\epsilon_{k, +}^*, \epsilon_{k, -}^*$ are non-negative variables. Note that $\epsilon_{k, +}^*, \epsilon_{k, -}^*$ are the largest uncertainty in the positive and negative direction, respectively, at instant $k$ given $x, u, f$, which results in non-zero contact forces (i.e., stability margin, see also \fig{fig:concept}).
%
%
\eq{bilevel_obj} calculates the smallest stability margin over time-horizons by subtracting the stability margin along the positive direction from that along the negative direction. 
Hence, we formulate a bilevel optimization problem which consists of two lower-level optimization problems as follows:
\begin{subequations}
\begin{flalign}
\max_{x, u, f, \epsilon_+^*, \epsilon_-^*} (\min _{k} \epsilon_{k, +}^* - \max _{k} -\epsilon_{k, -}^*)  \ \\
\text{s. t. } \quad \text{\eq{const2}, \eq{bounds_variables}}, \\
\epsilon_{k, +}^* \in \argmax_{\epsilon_{k, +}} \{\epsilon_{k, +}: A_k\epsilon_{k, +} \leq b_k , \epsilon_{k, +} \geq 0 \}, \label{bi-const1} \\
\epsilon_{k, -}^* \in \argmax_{\epsilon_{k, -}} \{\epsilon_{k, -}: -A_k\epsilon_{k, -} \leq b_k , \epsilon_{k, -} \geq 0 \}
\end{flalign}
\label{equation_sm_1}
\end{subequations}
where $A_k \in\mathbb{R}^{2 \times 1}, b_k \in\mathbb{R}^{2 \times1}$ represent inequality constraints in \eq{fna_cond_mass_one}, \eq{fnb_cond_mass}. 
 $A_k\epsilon_{k, +} \leq b_k , \epsilon_{k, +} \geq 0,$ and $-A_k\epsilon_{k, -} \leq b_k , \epsilon_{k, -} \geq 0$ represent the lower-level constraints for each lower-level optimization problem while \eq{const2}, \eq{bounds_variables} represent the upper-level constraints. $\epsilon_+, \epsilon_-$ are the lower-level objective functions while $\min _{k} \epsilon_{k, +}^* - \max _{k} -\epsilon_{k, -}^* $ is the upper-level objective function. $\epsilon_{k, +}, \epsilon_{k, -}$ are the lower-level decision variables of each lower-level optimization problem while $x, u, f, \epsilon_+^*, \epsilon_-^*$ are the upper-level decision variables. 


\eq{equation_sm_1} considers the largest one-side frictional stability margin along positive and negative direction at $k$. Therefore, by solving these two lower-level optimization problems, we are able to obtain the maximum frictional stability margin along a positive and negative direction. 
The advantage of \eq{equation_sm_1} is that since the lower-level optimization problem are formulated as two linear programming problems, we can efficiently solve the entire bilevel optimization problem using the Karush-Kuhn-Tucker (KKT) condition as follows:
\begin{subequations}
\begin{flalign}
 w_{k, +, j}, w_{k, -, j} \geq 0, C_k\epsilon_{k, +} \leq d_k , E_k\epsilon_{k, -} \leq d_k,\\
w_{k, +, j}(C_k\epsilon_{k, +} - d_k)_j = 0, \\
w_{k, -, j}(E_k\epsilon_{k, -} - d_k)_j = 0, \\
\nabla (-\epsilon_{k, +}) + \sum_{j=1}^{3}w_{k, +, j} \nabla (C_k\epsilon_{k, +} - d_k)_j= 0, \label{kkt1}\\
\nabla (-\epsilon_{k, -}) + \sum_{j=1}^{3}w_{k, -, j} \nabla (E_k\epsilon_{k, +} - d_k)_j= 0 \label{kkt2}
\end{flalign}
\label{kkt_equations}
\end{subequations}
where $C_k = [A_k^\top, -1]^\top \in\mathbb{R}^{3 \times 1}, d_k = [b_k^\top, 0]^\top \in\mathbb{R}^{3\times 1}, E_k = [-A_k^\top, -1]^\top \in\mathbb{R}^{3 \times 1}$.
$w_{k, +, j}$ is Lagrange multiplier associated with  $(C_k\epsilon_{k, +} \leq d_k)_j$, where $(C_k\epsilon_{k, +} \leq d_k)_j$ represents the $j$-th inequality constraints in $C_k\epsilon_{k, +} \leq d_k$. $w_{k, -, j}$ is Lagrange multiplier associated with  $(E_k\epsilon_{k, +} \leq d_k)_j$. 
$\nabla$ is the gradient operator with respect to $\epsilon_{k, +}$ in \eq{kkt1} and $\epsilon_{k, -}$ in \eq{kkt2}, respectively. 
Using the KKT condition and epigraph trick, we eventually obtain a single-level large-scale nonlinear programming problem with complementarity constraints: 
\begin{subequations}
\begin{flalign}
\max_{x, u, f, \epsilon_+^*, \epsilon_-^*, w_+, w_-} (t_+ + \alpha t_- ) \label{cost_bilevel}  \ \\
\text{s. t. } \quad \text{\eq{const2}, \eq{bounds_variables}, \eq{kkt_equations}},\\
t_+ \leq \epsilon_{k, +}^*, t_- \leq \epsilon_{k, -}^*, \forall k
\end{flalign}
\label{kkt_convertion}
\end{subequations}
where $\alpha$ is a weighting scalar. 
Note that we derive \eq{kkt_convertion} for the case with a uncertain mass parameter but this formulation can be easily converted to the case where uncertainty exists in CoM location by replacing $A_k, b_k$ in \eq{equation_sm_1} with \eq{fna_fnb_r}.  
Therefore, by solving tractable \eq{kkt_convertion}, we can efficiently generate robust trajectories that are robust against uncertain mass and CoM location parameters. 


\textit{Remark 1}:
If we consider the case where uncertainty exists in both mass and CoM location simultaneously, we would have a nonlinear coupling term $(C_x+r)(mg + \epsilon)$ in static equilibrium of moment. This makes the lower-level optimization non-convex optimization, making it extremely challenging to solve during bilevel optimization.

\section{Experimental Results}\label{sec:results}

In this section, we verify the performance of our proposed approach for pivoting. We present experiments to answer the following questions:
\begin{enumerate}
    \item How much robustness can our proposed bilevel optimization method provide over the baseline method?
    \item Can we demonstrate robustness of our proposed optimization during control of pivoting manipulation?
\end{enumerate}


\begin{table}[t]
    \caption{{Parameters of objects. $m, l, w$ represent the mass, length, and the width of the object, respectively. For pegs, the first element in $l, w$ are $l_1, w_1$ and the second element in $l, w$ are $l_2, w_2$, respectively, shown in \fig{fig:hardwareresults}.}}
    \centering
    \begin{tabular}{c|c|c|c|c}
     & $m$ [g] & $l$ [mm] & $w$ [mm] & $\mu_A, \mu_B, \mu_P$\\
         \hline\hline gear 1 & 140 & 84 & 20 & 0.3, 0.3, 0.8\\
         \hline gear 2 & 100 & 121 & 9.5 & 0.3, 0.3, 0.8\\
         \hline peg 1 & 45 & 36, 40 & 20, 28 & 0.3, 0.3, 0.8\\
         \hline peg 2 & 85 & 28, 40 & 10, 11 & 0.3, 0.3, 0.8
    \end{tabular}
    \label{parameter_table}
\end{table}

\begin{figure*}
    \centering
    \includegraphics[width=0.91\textwidth]{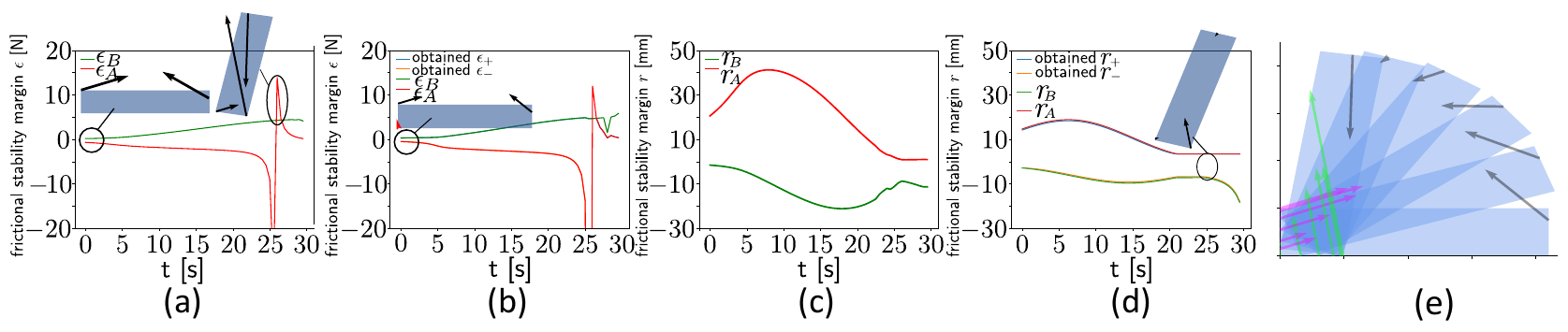} %
    \caption{Trajectory of frictional stability margin. $\epsilon_A, \epsilon_B$ are bounds of $\epsilon$ from \eq{fna_cond_mass_one}, \eq{fnb_cond_mass}. $r_A, r_B$ are bounds of $r$ from \eq{fna_fnb_r}.  $\epsilon_+, \epsilon_-, r_+, r_i$ are solutions obtained from the bilevel optimization. (a), (b): Trajectory of frictional stability of gear 1 based on uncertain mass obtained from baseline optimization, our proposed bilevel optimization, respectively. (c), (d): Trajectory of frictional stability of gear 1 based on uncertain CoM location obtained from baseline optimization, our proposed bilevel optimization, respectively. (e): Snapshots of pivoting motion for gear 1 obtained from our proposed bilevel optimization considering uncertain CoM location.}
    \label{fig:openloop_result}
\end{figure*}

\begin{figure}
    \centering
    \includegraphics[width=0.35\textwidth]{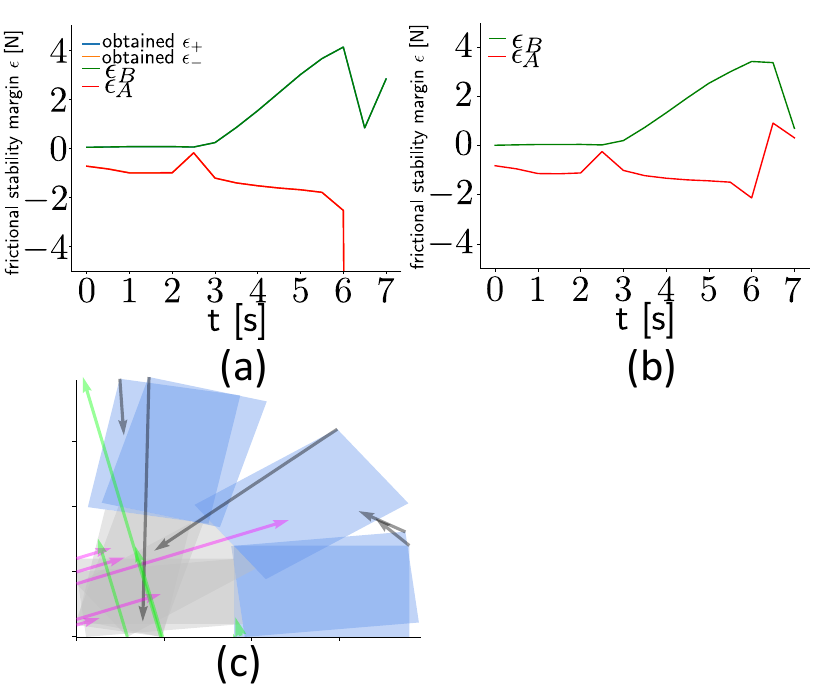} %
    \caption{(a), (b): Trajectory of frictional stability margin of peg 1 based on uncertain mass obtained from our proposed bilevel optimization, baseline optimization, respectively. (c): Snapshots of pivoting motion for peg 1, obtained from our proposed bilevel optimization considering uncertain mass.}
    \label{fig:openloop_result_peg1}
\end{figure}


\begin{table}[t]
    \caption{{Worst-case stability margin over the control horizon obtained from optimization for gear 1. Note that the stability margin for the solution of the benchmark optimization is analytically calculated.}}
    \centering
    \begin{tabular}{c|c|c}
     & $\epsilon_{+}^*$, $\epsilon_{-}^* $  [N]& $r_{+}^*$, $r_{-}^*$ [mm]\\
         \hline\hline Benchmark optimization \eq{equation_control} & 0.10, 0.66 & 1.5, 0.85\\
         \hline Ours  \eq{kkt_convertion} with mass uncertainty & 0.34, 0.50 & N/A \\
         \hline Ours  \eq{kkt_convertion} with CoM uncertainty & N/A & 3.43, 2.70
    \end{tabular}
    \label{epsilon}
\end{table}

\begin{table}[t]
    \caption{{Obtained worst stability margins over the time horizons from optimization for peg 1. Note that the stability margin for the solution of the benchmark optimization is analytically calculated.}}
    \centering
    \begin{tabular}{c|c|c}
     & $\epsilon_{+}^*$, $\epsilon_{-}^* $  [N]& $r_{+}^*$, $r_{-}^*$ [mm]\\
         \hline\hline Benchmark optimization \eq{equation_control} & 0.035, 0.018 & 31, 0\\
         \hline Ours  \eq{kkt_convertion} with mass uncertainty & 0.050, 0.021 & N/A \\
         \hline Ours  \eq{kkt_convertion} with CoM uncertainty & N/A & 38, 0
    \end{tabular}
    \label{epsilon_peg}
\end{table}

\subsection{Experiment Setup}
We implement our method in Python using IPOPT solver \cite{80fe29bf9dc245ffa5c8bd7b3eee2902} with PYROBOCOP \cite{DBLP:journals/corr/abs-2106-03220}. The optimization problem is implemented on a computer with Intel i7-8700K.

We demonstrate our algorithm on 4 different objects, as detailed in \tab{parameter_table}. During optimization, we set $Q=\text{diag}(0.1, 0), R=\text{diag}(0.01, 0.01)$. We use $N=60$ for gear 1, 2 and $N = 15$ for peg 1, 2, respectively. We use $\alpha= 0.001$ when we run \eq{kkt_convertion}. We set $x_s = [0, \frac{w}{4}]^\top, \theta_g = \frac{\pi}{2}$. 


For hardware experiments, we use a Mitsubishi Electric Assista industrial manipulator arm (see \fig{fig:pivoting_abstractfig}). We use a force controller which is designed using the default stiffness controller of the robot. We test our method on 4 different objects listed in \tab{parameter_table}.

\subsection{Results of Numerical  Optimization}
\fig{fig:openloop_result} shows the time history of frictional stability margin of gear 1 obtained from our proposed bilevel optimization considering uncertain mass and uncertain CoM location, and the benchmark optimization. Overall, our proposed optimization could generate more robust trajectories. 
For example, at $t=0$ s, $f_{nB}$ in (a) is almost zero so that the stability margin obtained from \eq{fnb_cond_mass} is almost zero. In contrast,  our proposed optimization could realize non-zero $f_{nB}$ as shown as a red arrow in (b). 
In (d), to increase the stability margin, the finger position $p_y$ moves on the face of gear 1 so that the controller can increase the stability margin more than the benchmark optimization. This would not happen if we do not consider complementarity constraints \eq{slippingP}. 
Also, our obtained $\epsilon_{+}, \epsilon_{-}, r_+, r_-$ follows bounds of stability margin. It means that our proposed bilevel optimization can successfully design a controller that maximizes the worst stability margin given the best stability margin for each time-step.  
%


\tab{epsilon} and \tab{epsilon_peg} summarize the computed stability margin from \fig{fig:openloop_result}. In \tab{epsilon}, for the case where our optimization considers uncertainty of mass, we observe that the value of $\epsilon_-^*$ from our optimization is smaller than that from the benchmark optimization although the sum of the stability margin $\epsilon_+^* + \epsilon_-^* $ from our bilevel optimization is greater than that from the benchmark optimization. This result means that our optimization can actually improve the worst-case performance by sacrificing the general performance of the controller. Regarding the case where we consider the uncertain CoM location, our proposed bilevel optimization outperforms the benchmark trajecoty optimization in both $r_+^*, r_-^*$. For peg 1, the bilevel optimizer finds trajectories that have larger stability margins for both uncertain mass and CoM location as shown in \tab{epsilon_peg}. The trajectory of stability margin obtained from bilevel optimization considering mass uncertainty is illustrated in \fig{fig:openloop_result_peg1}.

\subsection{Hardware Experiments}
We implement our controller using a 6 DoF manipulator to demonstrate the efficacy of our proposed method for gear 1. 
 To evaluate robustness for objects with unknown mass, we solve the optimization with mass different from the true mass of the object and implement the obtained trajectory on the object. 
 We implement trajectories obtained from the two different optimization techniques using 4 different mass values.

\begin{table}[t]
    \caption{{Number of successful pivoting attempts of gear 1 over 10 trials for the two different methods. To evaluate robustness for objects with unknown mass, we solve the optimization with  mass different from the known object and implement the obtained trajectory on the object with known mass. Note that the actual mass of gear 1 is 140 g. }}
    \centering
    \begin{tabular}{c|c|c}
 & Bilevel Optimization & Benchmark Optimization\\
         \hline\hline $m=100$ g  & 10 / 10 & 0 / 10 \\
         \hline $m=110$ g  & 10 / 10 & 0 / 10 \\
         \hline $m=140$ g  & 10 / 10 & 0 / 10 \\
         \hline $m=170$ g  & 10 / 10 & 0 / 10 
    \end{tabular}
    \label{hardware_result}
\end{table}

\begin{figure}
    \centering
    \includegraphics[width=0.465\textwidth]{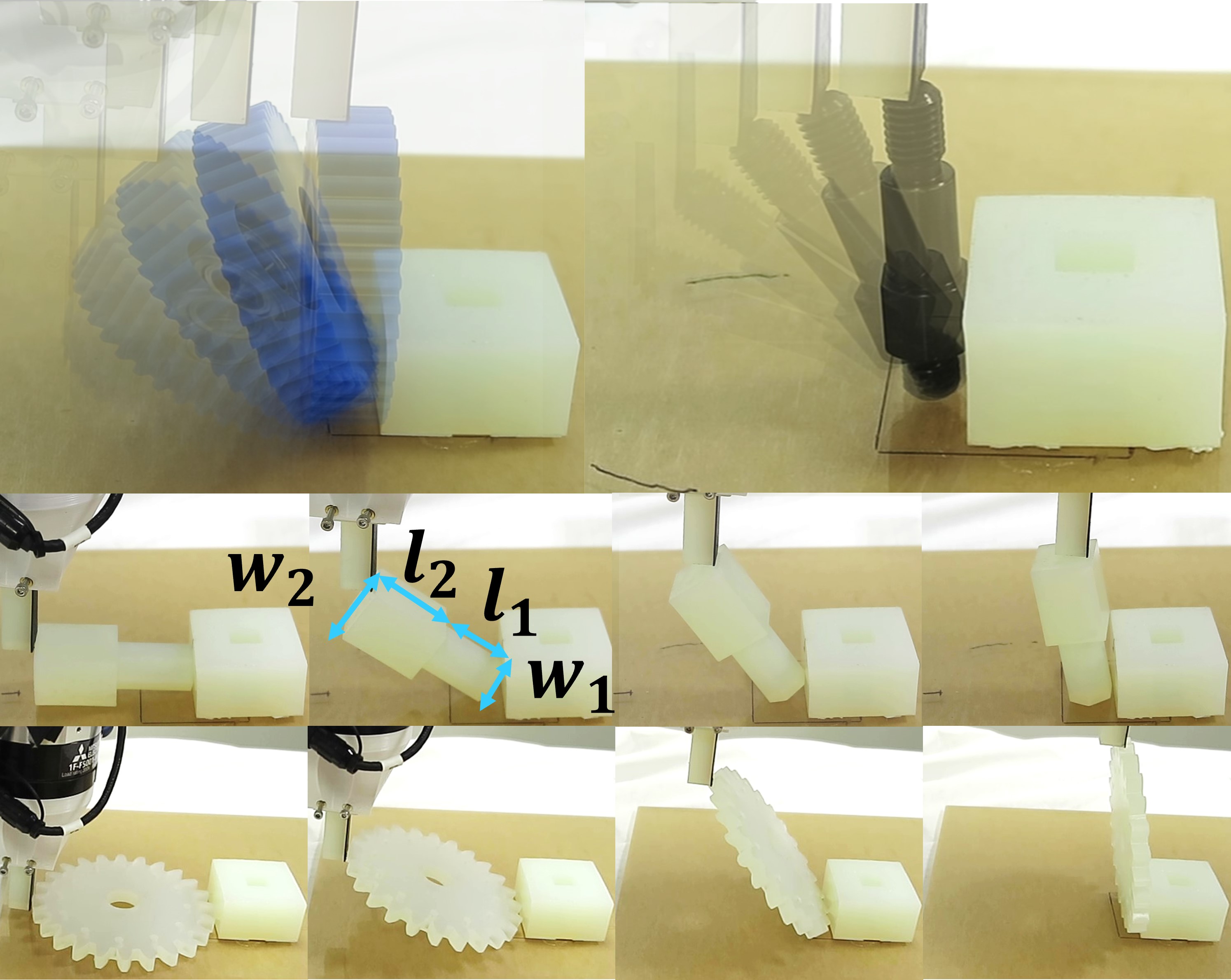} %
    \caption{Snapshots of hardware experiments. We show snapshots of the white peg and gear (instead of overlaid images) for clarity.}
    \label{fig:hardwareresults}
\end{figure}

\tab{hardware_result} shows the success rate of pivoting for the hardware experiments. We observe that our proposed bilevel optimization is able to achieve 100 $\%$ success rates for all $4$ mass values while benchmark optimization cannot realize stable pivoting. 
Note that the benchmark trajectory optimization also generates trajectories with non-zero frictional stability margins but they failed to pivot the object. The reason would be that the system has a number of uncertainties such as incorrect coefficient of friction, sensor noise in the F/T sensor (for implementing the force controller), etc. which are not considered in the model. We believe that these uncertainties make the objects unstable leading to failure of pivoting.  In contrast, even though our proposed bilevel optimization also does not consider these uncertainties, it generates more robust trajectories and we believe that this additional robustness could account for the unknown uncertainty in the real hardware.  We also observe that the trajectories generated by benchmark optimization can successfully realize pivoting if the manipulator uses patch contact during manipulation (thus getting more stability).
Finally, \fig{fig:hardwareresults} shows the snapshots of hardware experiments for the 4 objects detailed in \tab{parameter_table}. We observe that our bilevel optimization can successfully pivot all 4 objects during hardware experiments.

\section{Discussion and Future Work}\label{sec:discussion}

This paper presents \textit{frictional stability}-aware optimization, a strategy that exploits friction for robust planning of pivoting. By considering uncertainty in mass and CoM location, we discussed the stability margin for slipping contact. We presented our proposed tractable robust bilevel optimization formulation. We verified the robustness of our proposed method for pivoting using simulation and hardware experiments. 


In the future, we will try to understand the following questions: 

\textbf{Frictional Stability of Patch Contact}: During the hardware experiments, we observed that patch contact provides additional robustness with the manipulation. We would presume that modeling patch contacts in our proposed framework would expand frictional stability margin. 

\textbf{General Frictional Stability}: This work assumes that uncertainty arises from either mass or CoM location. We would also like to consider other uncertainties such as coefficient of friction, kinematics uncertainty, etc. However, these uncertainties lead to stochastic complementarity system, which requires the discussion in e.g., \cite{drnach2021robust, yuki2022chance}. Also, solving our proposed optimization can be intractable once we consider a nonlinear coupling term as described in Remark 1. Thus, it is also important to formulate computationally tractable optimization problems to solve more general frictional stability problems. 

\textbf{Feedback Control using Frictional Stability}: The proposed method was implemented in an open-loop fashion. We would argue that the frictional stability would increase with active feedback control for compensating for pose errors or error recovery using tactile sensors e.g., \cite{donlon2018gelslim}.

\bibliographystyle{IEEEtran}
\bibliography{main.bib}

\end{document}